\documentclass[final,3p,times]{elsarticle}
\pdfoutput=1  
\usepackage{setspace}
\usepackage{csquotes}
\usepackage{bm}
\usepackage{booktabs}
\usepackage{bigstrut}
\usepackage{multirow}
\usepackage[short]{optidef}
\usepackage{enumerate}
\usepackage{subcaption}
\usepackage{adjustbox}
\usepackage[table,xcdraw]{xcolor}
\usepackage{algorithm}
\usepackage{amssymb}

\usepackage[noend]{algpseudocode}
\usepackage[capitalise]{cleveref}
\usepackage{url}
\usepackage{csvsimple}
\usepackage{epstopdf}
\usepackage{float}
\usepackage{url}
\usepackage{footnote}
\usepackage[many]{tcolorbox}
\DeclareUrlCommand\ULurl{}
\usepackage{adjustbox}
\mathchardef\mhyphen="2D % Define a "math hyphen"

\begin{document}

\begin{frontmatter}


\author[label1]{Chandan Gautam\corref{cor1}}
\ead{chandangautam31@gmail.com phd1501101001@iiti.ac.in}
\author[label1]{Aruna Tiwari}
\ead{artiwari@iiti.ac.in}
\author[label2]{M. Tanveer}
\ead{mtanveer@iiti.ac.in}

\cortext[cor1]{Corresponding author}
\address[label1]{Discipline of Computer Science and Engineering, IIT Indore, Simrol, India }
\address[label2]{Discipline of Mathematics, IIT Indore, Simrol, India }
%\fntext[label2]{School of Electrical and Electronic Engineering, Nanyang Technological University Singapore}

\title{}
\title{OCKELM+: \textcolor{red}{Kernel Extreme Learning Machine} based One-class Classification using Privileged Information \\or\\
(KOC+: \textcolor{red}{Kernel Ridge Regression} or \textcolor{red}{Least Square SVM with zero bias} based One-class Classification using Privileged Information)}

\address{}

\begin{tcolorbox}[width=7in,colback=white]
	\textcolor{red}{Why three names in the paper title:} \textcolor{blue}{Because three methods viz; Kernel ridge regression (KRR), lease square support vector machine with zero bias (LSSVM(bias=0)) and kernel extreme learning machine (KELM), are identical in outcomes and developed by three different researchers under three different framework. Since, KRR are more genric name compared to others, we used name KRR instead of LSSVM or KELM in this paper. Proposed methods of this paper can be considered as variants of KRR or LSSVM(with bias=0) or KELM:}
	
	\centering {\textcolor{red}{\textbf{KELM = KRR = LSSVM(with bias=0)}}}
\end{tcolorbox}

\begin{abstract}
	Kernel method-based one-class classifier is mainly used for outlier or novelty detection. In this letter, kernel ridge regression (KRR) based one-class classifier (KOC) has been extended for learning using privileged information (LUPI). LUPI-based KOC method is referred to as KOC+. This privileged information is available as a feature with the dataset but only for training (not for testing). KOC+ utilizes the privileged information differently compared to normal feature information by using a so-called correction function. Privileged information helps KOC+ in achieving better generalization performance which is exhibited in this letter by testing the classifiers with and without privileged information. Existing and proposed classifiers are evaluated on the datasets from UCI machine learning repository and also on MNIST dataset. Moreover, experimental results evince the advantage of KOC+ over KOC and support vector machine (SVM) based one-class classifiers.            
\end{abstract}

\end{frontmatter}

\section{Introduction}\label{sec:Intro}
In recent years, learning using privileged information ($LUPI$) framework is being quite popular among researchers \cite{wang2018learning,chevalier2018classifying,lambert2018deep,smolyakov2016one,motiian2016information,zhang2015support,vapnik2015learning,lapin2014learning,zhu2014new,feyereisl2012privileged}. In the real world, human learns not just by looking at an object but also learns by listening to extra information provided by someone. For an example, a student learns a concept not only by the available  explanation in the book but also learns by listening comments of the teacher. Here, comments are privileged information. Vapnik and Vashist \cite{vapnik2009new} inspired by this student-teacher learning concept and proposed a $LUPI$ framework which explores privileged information to improve the performance of the classifier. Further, this concept has been explored for various types of tasks viz., face verification \cite{xu2015distance}, multi-Label classification \cite{wang2018learning}, visual recognition \cite{motiian2016information}, malware detection \cite{smolyakov2016one} etc. Most recently, $LUPI$ framework has been employed for one-class classification ($OCC$) task \cite{smolyakov2016one,zhang2015support,zhu2014new}. We are also going to explore $LUPI$ for $OCC$ task in this paper.

One-class classification using kernel learning is a well-known approach for outlier or novelty detection. Support vector machine (SVM) based methods viz., one-class SVM (OCSVM)\cite{scholkopf1999support} and support vector data description (SVDD)\cite{tax1999support}, are the most popular among kernel-based methods. In the recent years, kernel ridge regression (KRR) based binary or multi-class classifiers have received quite attention by the researchers due to its no-iterative approach of learning ability. Most recently, KRR-based one-class classifier (KOC)\footnote{\label{koc_elm}In the paper\cite{leng2014one}, one-class classifier is developed based on kernel extreme learning machine and named as one-class extreme learning machine (OCELM). However, we didn't find any differences in the formulation and outcomes of kernel extreme learning machine and KRR. Therefore, we have decided to follow the generic name KRR instead of extreme learning machine and named as KOC instead of OCELM.} has been developed by the Leng et al.\cite{leng2014one}\textsuperscript{\ref{koc_elm}}. This method, KOC, can also be think as the variant of least square SVM (with bias=0) or, variant of kernelized extreme learning machine.  In the recent advancement of kernel learning, performance of kernel learning-based methods has been improved by introducing a new learning paradigm i.e.learning using privileged information (LUPI)\cite{vapnik2009new}. They have introduced group attributes as a privileged information with their proposed optimization problem and experimentally verified that these privileged information has significantly improved the performance. LUPI concept is inspired from student-teacher learning among human being where a student learns from the explanation and comments of his teacher. Further, this concept has been employed with OCSVM and SVDD, and created one-class classifiers OCSVM+\cite{zhu2014new} and SVDD+\cite{zhang2015support}, respectively. Most recently, Burnaev and Smolyakov\cite{smolyakov2016one} modified the formulation of OCSVM+ and SVDD+ by adding a regularization factor on privileged feature space. OCSVM+ and SVDD+ have yielded significant improvement over tradition OCSVM and SVDD. In this letter, KOC is enabled to utilize LUPI framework with it and proposed an one-class classifier i.e. KOC+. 

The rest of the paper is organized as follows. We first discuss standard KOC in sections \ref{sec:KOC} as it is the base method for the proposed method KOC+. Section \ref{sec:Priv_defin+} describes privileged information and privileged information-based two existing SVM-based one-class classifiers viz., OCSVM+ and SVDD+. Further, the proposed LUPI framework-based KOC (i.e. KOC+) is presented in section \ref{sec:KOC+}. Section \ref{sec:exp} describes the experimental setup, and evaluates the proposed (KOC+) and existing One-class Classifiers (KOC, OCSVM+, and SVDD+) against datasets from UCI repository \cite{Dua:2017} and on MNIST dataset. The paper concludes in Section \ref{sec:concl}.

\section{KRR-based One-class Classifier (KOC)}\label{sec:KOC}
Let us assume the input training matrix of size $N$ is $\bm{X=\left\{x_i\right\}}$, where $\bm{x_i}=[x_{i1}, x_{i2},...,x_{in}]$, $i=1,2,...,N$, is the $n$-dimensional input vector of the $i^{th}$ training sample. KOC\cite{leng2014one}\textsuperscript{\ref{koc_elm}} approximates all samples to any real number r and creates a boundary for the classifier. The minimization function of KOC is written as follows:
\begin{equation}\label{Eq:KOC}
\begin{aligned}
\underset{\bm{\beta},e_i}{\text{Min}}:\pounds_{KOC}=\frac{1}{2}\left \|\bm{\beta}\right\|^{2} + C\frac{1}{2}\sum_{i=1}^{N}\left\|e_i  \right \|_2^{2}  \\
\text{Subject to}:\ \bm{\phi_i \cdot \beta}=r - e_i, \text{  }i=1,2,...,N
\end{aligned}
\end{equation}

where $\bm{\beta}$ denotes weight matrix for KRR-based classifiers, $\phi(.)$ denotes kernel feature mapping function, $\bm{\phi_i =\phi(x_i)}$, and $\bm{\Phi}= \bm{\Phi(X)} = \left [\bm{\phi_1, \phi_2,...,\phi_N\right ]}$. $\bm{E}$ is an error vector where $\bm{E}=\left\{e_i\right\}$, where $i=1, 2,...,N$. Here, $\bm{r}$ is a vector having all elements equal to $r$ and $r$ is any real number but we assume $r=1$ as data from only one-class is available.

After solving the minimization problem in (\ref{Eq:KOC}), weight matrix is obtained as follows:

\begin{equation}
\label{Eq:ow_KOC}
\begin{aligned}
\bm{\beta} = \bm{\Phi}\left(\bm{K}+\frac{1}{C}\bm{I}\right)^{-1}\bm{r}
\end{aligned}
\end{equation}

where  Kernel matrix is denoted as $\bm{K}=\bm{\Phi \cdot \Phi}$. Further, decision function for predicting whether a testing sample is an outlier or not is same as discussed in Section \ref{dec_fun}.

%\begin{align}
%  Q=\left(N+\frac{1}{2}\right)e
%\end{align}

\section{Privileged Information}\label{sec:Priv_defin+}

A teacher doesn't play any significant role in the traditional algorithms of machine learning. However, human learns not just by looking at an object but also learns by listening to extra information provided by teachers like comments, explanations etc. Here, comments and explanations can be treated as a privileged information\cite{vapnik2015learning}. We can understand the importance of this concept by following examples:

\begin{enumerate}[(i)]
	
\item Suppose our goal is to build a model which can predict outcome of
a treatment in a year based on the currently visible symptoms of a patient. However, it is possible to provide some additional information like, the development of symptoms in three months, in six months, and in nine months, at the training stage\cite{vapnik2009new}. It is to be noted that these additional information can not be available during testing, and, therefore, these additional information can be used as privileged information to improve the model prediction. 

\item Suppose we have to build a model to classify biopsy images into two categories: cancer and non-cancer. Here, model classifies based on the pixel information of image, however, an additional information can be provided along with the image\cite{vapnik2009new}. This additional information can be a description about image which is written by a pathologist. This description provides additional information, i.e., description of the pictures using a high level holistic language. However, this additional information will not be available at the test stage because our goal is to build a model to provide an an accurate diagnosis without consulting with a pathologist\cite{vapnik2009new}.

\item  Generally, images are directly feed to the machine without any extra information. However, it is possible that we can provide some extra information\cite{sharmanska2013learning} like, attribute annotation, bounding box annotation, textual description and rationales, by the help of human during training. Sharmanska et al.\cite{sharmanska2013learning} treated these information as a privileged information and provide these extra information to the machine during training. During testing, these information can not be available as our goal is to built a model which can predict correct class of testing image without any external help. 
\end{enumerate}    

\subsection{LUPI framework with OCSVM: OCSVM+}\label{ocsvm+}
Let us assume the input training matrix of size $N$ is $\bm{X=\left\{x_i\right\}}$, where $\bm{x_i}=[x_{i1}, x_{i2},...,x_{in}]$, $i=1,2,...,N$, is the $n$-dimensional input vector of the $i^{th}$ training sample. We assume that privileged information $\bm{X^*=\left\{x_i^*\right\}}$, where $i=1, 2,...,N$, is available with feature space of $X$ as $(\bm{x_i,x_i^*})$ during training. OCSVM constructs a hyper-plane such that it separates all the data points from the origin in such a way so that the distance of hyper-plane
from the origin is maximum. The minimization function of OCSVM+ is as follows\cite{smolyakov2016one,zhu2014new}:

%\begin{mini}
%	{\bm{\omega} ,\bm{\xi}, \rho, \bm{\omega^*}, b^*}{\dfrac{\nu N}{2}\|\bm{\omega}\|_2^{2} + \frac{\mu}{2}\left \|\bm{\omega^*}\right\|^{2} - \nu N \rho + \sum_{i=1}^{N}[(\bm{\omega^*}. \bm{\phi^* (x_i^*)}) + b^* +\xi_i]}{}{}
%	\addConstraint{(\bm{\omega}. \bm{\phi (x_i)}) \geq \rho - (\bm{\omega^*}. \bm{\phi^* (x_i^*)}) - b^*, \quad}{i=0,\ldots,N}
%	\addConstraint{(\bm{\omega^*}. \bm{\phi^* (x_i^*)}) + b^* + \xi_i \geq 0, \quad}{i=0,\ldots,N}
%	\addConstraint{\xi_i\geq 0,\quad} {i=0,\ldots,N}
%	\label{eq:ocsvm+_primal}
%\end{mini}

\begin{equation}\label{Eq:ocsvm+_primal}
\begin{aligned}
	\underset{\bm{\omega} ,\bm{\xi}, \rho, \bm{\omega^*}, b^*}{\text{Min}}:  \pounds_{OCSVM+}={\dfrac{\nu N}{2}\|\bm{\omega}\|_2^{2} + \frac{\mu}{2}\left \|\bm{\omega^*}\right\|^{2} - \nu N \rho + \sum_{i=1}^{N}[(\bm{\omega^*}. \bm{\phi^* (x_i^*)}) + b^* +\xi_i]}{}{}\\
	\text{Subject to}:\ {(\bm{\omega}. \bm{\phi (x_i)}) \geq \rho - (\bm{\omega^*}. \bm{\phi^* (x_i^*)}) - b^*, \quad} {i=0,\ldots,N}\\
	{(\bm{\omega^*}. \bm{\phi^* (x_i^*)}) + b^* + \xi_i \geq 0, \quad}{i=0,\ldots,N}\\
	{\xi_i\geq 0,\quad} {i=0,\ldots,N}
\end{aligned}
\end{equation}

where $\bm{\omega}$ denotes weight matrix\footnote{We have used two different notations for weight matrix of SVM and KRR to avoid confusion.}, and $\bm{\omega^*}$ is a correction weight for SVM-based classifiers. $\phi(.)$ denotes kernel feature mapping function, and $\bm{\phi^*(.)}$ is a feature mapping in the privileged space. $\mu$ is a regularization parameter, and $\xi_i$ is a slack variable for the $i^{th}$ pattern. $\rho$ and $b^*$ are the bias terms.

\subsection{LUPI framework with SVDD: SVDD+}\label{svdd+}
SVDD constructs a hyper-sphere of minimum radius around the target class data in such a way so that it encloses almost all points in the target class data set. LUPI framework for SVDD is developed as follows\cite{smolyakov2016one,zhang2015support}:
%\begin{mini}
%	{\bm{\xi}, \bm{\omega^*}, b^*, \bm{a}, R}{\nu N R + \frac{\mu}{2}\left \|\bm{\omega^*}\right\|^{2} + \sum_{i=1}^{N}[(\bm{\omega^*}. \bm{\phi^* (x_i^*)}) + b^* +\xi_i]}{}{}
%	\addConstraint{\|\bm{\phi(x_i)}-\bm{a}\|^2 \leq  R + [(\bm{\omega}. \bm{\phi^* (x_i^*)}) + b^*], \quad}{i=1,\ldots,N}
%	\addConstraint{(\bm{\omega^*}. \bm{\phi^* (x_i^*)}) + b^* + \xi_i \geq 0, \quad}{i=0,\ldots,N}
%	\addConstraint{\xi_i\geq 0,\quad} {i=0,\ldots,N}
%	\label{eq:svdd+_primal}
%\end{mini}

\begin{equation}\label{eq:svdd+_primal}
\begin{aligned}
\underset{\bm{\xi}, \bm{\omega^*}, b^*, \bm{a}, R}{\text{Min}}:  \pounds_{SVDD+}={\nu N R + \frac{\mu}{2}\left \|\bm{\omega^*}\right\|^{2} + \sum_{i=1}^{N}[(\bm{\omega^*}. \bm{\phi^* (x_i^*)}) + b^* +\xi_i]}{}{}\\
\text{Subject to}:\ {\|\bm{\phi(x_i)}-\bm{a}\|^2 \leq  R + [(\bm{\omega}. \bm{\phi^* (x_i^*)}) + b^*], \quad}{i=1,\ldots,N}\\
 {(\bm{\omega^*}. \bm{\phi^* (x_i^*)}) + b^* + \xi_i \geq 0, \quad}{i=0,\ldots,N}\\
 {\xi_i\geq 0,\quad} {i=0,\ldots,N}
\end{aligned}
\end{equation}

where $\bm{a}$ is a center, and $R$ is a radius of the hypersphere. Rest notations are same as discussed for OCSVM+.

\section{Proposed Method: LUPI framework with KOC (KOC+)}\label{sec:KOC+}
In this section, optimization problem of KOC is modified according to LUPI framework and provides a minimization problem for KOC+. Let us do some assumption as per LUPI framework. We assume that privileged information $\bm{X^*=\left\{x_i^*\right\}}$, where $i=1, 2,...,N$, is available with feature space of $X$ as $(\bm{x_i,x_i^*})$ during training. However, this information is not available during testing. As suggested in \cite{vapnik2009new}, privileged information is incorporated to the optimization problem by modeling the slack variable $e_i$ as so called correction function:

\begin{equation}\label{Eq:corr_func}
\begin{aligned}
e_i = \bm{e_i(x_i^*)} = \bm{\phi^*(x_i^*). \beta^*} = \bm{\phi_i^*. \beta^*}
\end{aligned}
\end{equation}
where $\bm{\phi^*(.)}$ is a feature mapping in the privileged space and $\bm{\beta^*}$ is a correction weight. Now, we substitute (\ref{Eq:corr_func}) into (\ref{Eq:KOC}) and modify the optimization problem for KOC+ as follows:           
\begin{equation}\label{Eq:KOC+}
\begin{aligned}
\underset{\bm{\beta},\bm{\beta^*},e_i}{\text{Min}}:\pounds_{KOC+}=\frac{1}{2}\left \|\bm{\beta}\right\|^{2} + \mu\frac{1}{2}\left \|\bm{\beta^*}\right\|^{2} + C\frac{1}{2}\sum_{i=1}^{N}(\bm{\phi_i^* \cdot \beta^*})^{2}  \\
\text{Subject to}:\ \bm{\phi_i \cdot \beta}=r - \bm{\phi_i^* \cdot \beta^*}, \text{  }i=1,2,...,N
\end{aligned}
\end{equation}
where, $\left \|\bm{\beta}\right\|^{2}$ reflects the capacity of the decision function and $\left \|\bm{\beta^*}\right\|^{2}$ reflects the capacity of the correction function. Here, $\mu$ controls the capacity of these two functions i.e. controls the relative weight of these two capacities.

By using Representer Theorem \cite{argyriou2009there}, $\bm{\beta}$ can be expressed as a linear combination of the training data representation in non-linear feature space $\bm{\Phi}$ and reconstruction weight vector $\bm{W}$:
\begin{equation}
\label{Eq:representation_theorem_oc}
\begin{aligned}
\bm{\beta} = \bm{\Phi \cdot W} \quad \text{ and } \quad
\bm{\beta*} = \bm{\Phi^* \cdot W^*}.
\end{aligned}
\end{equation}

By substituting the (\ref{Eq:representation_theorem_oc}) into (\ref{Eq:KOC+}), following minimization problem is obtained:
\begin{equation}\label{Eq:KOC+_phi}
\begin{aligned}
\underset{\bm{W},\bm{W^*},e_i}{\text{Min}}:\pounds_{KOC+}=\frac{1}{2}\bm{\Phi \cdot \Phi} \cdot \left \|\bm{W}\right\|^{2} + \mu\frac{1}{2}\bm{\Phi^* \cdot \Phi^*} \cdot\left \|\bm{W^*}\right\|^{2} + \\ C\frac{1}{2}\sum_{i=1}^{N}(\bm{\phi_i^* \cdot \phi_i^* \cdot W^*})^{2}  \\
\text{Subject to}:\ \bm{\phi_i \cdot \phi_i \cdot W}=r - \bm{\phi_i^* \cdot \phi_i^* \cdot W^*}, \text{  }i=1,2,...,N
\end{aligned}
\end{equation}

Now we substitute $\bm{K}=\bm{\Phi \cdot \Phi}$ and $\bm{K^*}=\bm{\Phi^* \cdot \Phi^*}$ in (\ref{Eq:KOC+_phi}) then (\ref{Eq:KOC+_phi}) can be reformulated as follows:
\begin{equation}\label{Eq:KOC+_kern}
\begin{aligned}
\underset{\bm{W},\bm{W^*},e_i}{\text{Min}}:\pounds_{KOC+}=\frac{1}{2}\bm{K} \cdot \left \|\bm{W}\right\|^{2} + \mu\frac{1}{2}\bm{K^*} \cdot\left \|\bm{W^*}\right\|^{2} + \\C\frac{1}{2}\sum_{i=1}^{N}(\bm{k_i^* \cdot W^*})^{2}  \\
\text{Subject to}:\ \bm{k_i \cdot W}=r - \bm{k_i^* \cdot W^*}, \text{  }i=1,2,...,N
\end{aligned}
\end{equation}

where $\bm{k_i}\subseteq \bm{K}$ and $\bm{k_i^*}\subseteq \bm{K^*}$.

The Lagrangian relaxation of (\ref{Eq:KOC+_kern}) can be written as follows:
\begin{equation}\label{Eq:KOC+_kern2}
\begin{aligned}
&\underset{\alpha_i}{\text{Max}}\quad \underset{\bm{W},\bm{W^*},e_i}{\text{Min}}:\pounds_{KOC+} = \frac{1}{2}\bm{K} \cdot \left \|\bm{W}\right\|^{2} + \mu\frac{1}{2}\bm{K^*} \cdot\left \|\bm{W^*}\right\|^{2} + \\ &  C\frac{1}{2}\sum_{i=1}^{N}(\bm{k_i^* \cdot W^*})^{2} - \sum_{i=1}^{N}\bm{\alpha_i}(\bm{k_i \cdot W} - r + \bm{k_i^* \cdot W^*}),\\& \text{  }i=1,2,...,N
\end{aligned}
\end{equation}

where $\bm{\alpha = \{\alpha_i\}}, i=1,2 \hdots N$, is a Lagrangian multiplier. In order to optimize (\ref{Eq:KOC+_kern2}), set the partial derivatives of (\ref{Eq:KOC+_kern2}) is equal to zero as follows:
\begin{equation}
\label{Eq:KOC+_deriv1}
\begin{aligned}
&\frac{\partial \pounds_{KOC+}}{\partial \bm{W}} = 0 \Rightarrow \bm{W}=\bm{\alpha}
\end{aligned}
\end{equation}
\begin{equation}
\label{Eq:KOC+_deriv2}
\begin{aligned}
&\frac{\partial \pounds_{KOC+}}{\partial \bm{W^*}} = 0 \Rightarrow \mu\bm{K^*\cdot W^*} + C\bm{K^*\cdot W^* \cdot K^*} -\bm{\alpha \cdot K^*}=0
\end{aligned}
\end{equation}
\begin{equation}
\label{Eq:KOC+_deriv3}
\begin{aligned}
&\frac{\partial \pounds_{KOC+}}{\partial \bm{\alpha}} = 0 \Rightarrow \bm{K \cdot W + K^* \cdot W^* = r} \\
\end{aligned}
\end{equation}
Now, substitute (\ref{Eq:KOC+_deriv1}) and (\ref{Eq:KOC+_deriv3}) into (\ref{Eq:KOC+_deriv2}):
\begin{equation}
\label{Eq:KOC+_output1}
\begin{aligned}
\mu(\bm{r}-\bm{K \cdot W}) + C(\bm{r}-\bm{K \cdot W})\cdot \bm{K^*} - \bm{W} \cdot \bm{K^*} = 0
\end{aligned}
\end{equation}
 
After solving the (\ref{Eq:KOC+_output1}), $W$ is obtained as follows:
\begin{equation}
\label{Eq:KOC+_ow}
\begin{aligned}
\bm{W} = (\mu\bm{K}+C\bm{K}\cdot \bm{K^*}+\bm{K^*})^{-1} \cdot (\mu\bm{I}+C\bm{K^*})\cdot \bm{r}
 \end{aligned}
 \end{equation}
 The weight $\bm{\beta}$ is obtained by substituting (\ref{Eq:KOC+_ow}) into (\ref{Eq:representation_theorem_oc}) as follows:
 \begin{equation}
 \label{Eq:KOC+_beta}
 \begin{aligned}
 \bm{\beta} = \bm{\Phi}\cdot(\mu\bm{K}+C\bm{K}\cdot \bm{K^*}+\bm{K^*})^{-1} \cdot (\mu\bm{I}+C\bm{K^*})\cdot \bm{r}
 \end{aligned}
 \end{equation}
 
 The predicted output of KOC+ for training samples can be calculated as follows: %$\bm{\widehat{O}}= \bm{\Phi.\beta}=\bm{\Phi.\Phi.W}=\bm{K.W}$,
 
 \begin{equation}
 \label{Eq:out_koc+}
 \begin{aligned}
 \bm{\widehat{O}}= \bm{\Phi.\beta}=\bm{\Phi.\Phi.W}=\bm{K.W}
 \end{aligned}
 \end{equation}
 where $\bm{\widehat{O}}=\left\{\widehat{O}_i\right\}$, and $i=1, 2,...,N$, is the predicted output for training data.
 
\subsection{Decision Function}\label{dec_fun}
Further, a threshold ($\theta$) is employed with the proposed method, which is determined as follows:

\begin{enumerate}[(i)]
	\item Calculate distance between the predicted value of the $i^{th}$ training sample and $r$, and store in a vector, $\bm{d}=\left\{d_i\right\}$ and $i=1, 2,...,N$, as follows:
	\begin{equation}
	\label{Eqaekoc:dist1}
	\begin{aligned}	
	d_i=\left |\widehat{O}_i - r\right |
	\end{aligned}
	\end{equation}
	\item After storing all distances in $\bm{d}$ as per (\ref{Eqaekoc:dist1}), sort these distances in decreasing order and denoted by a vector $\bm{d_{dec}}$. Further, reject few percents of training samples based on the deviation. Most deviated samples are rejected first because they are most probably far from the distribution of the target data. The threshold is decided based on these deviations as follows:
	\begin{equation}
	\label{Eqaekoc:Thr1}
	\begin{aligned}	
	\theta=d_{dec}(\left \lfloor{\text{$\eta*N$}}\right \rfloor)\\
	\end{aligned}
	\end{equation}
	where $0<\eta\leq1$ is the fraction of rejection of training samples for deciding threshold value. $N$ is the number of training samples and $\lfloor\text{ } \rfloor$ denotes floor operation.
\end{enumerate}

After determining a threshold value by the above procedure, during testing, a test vector $\bm{{x}_p}$ is fed to the trained architecture and its output $\widehat{O}_p$ is obtained. Further, compute the distance ($\widehat{d_p}$), for $\bm{{x}_p}$, between the predicted value $\widehat{O}_p$ of the $p^{th}$ testing sample and $r$:
\begin{equation}
\label{Eqaekoc:dist1_1}
\begin{aligned}	
\widehat{d_p}=\left |\widehat{O}_p - r\right |
\end{aligned}
\end{equation}

Finally, $\bm{{x}_p}$ is classified based on the following rule:

\begin{equation}
\label{Eqaekoc:df_ocelm2}
\begin{aligned}
&\text{If } \widehat{d_p} \leq \text{Threshold}, &\text{$\bm{{x}_p}$ belongs to normal class} \\
&\text{Otherwise}, &\text{$\bm{{x}_p}$ is an outlier}
\end{aligned}
\end{equation}

\section{Experiments}\label{sec:exp}
In our experiments, we have tested on 7 datasets which were generated using 3 multi-class datasets viz., MNIST, Statlog Heart, and Abalone. These datasets are binary or multi-class datasets. We make these datasets compatible with one-class by assuming alternately, one class as a target class and remaining classes as outlier class. The Gaussian kernel is used for all experiments. Gaussian kernel parameter is selected from the range of $20$ values between $10^{-10}$ and $10^{20}$ with an equal interval. Another parameter $\eta$ is selected from the range of $10$ values between $0.05$ and $0.7$ with an equal interval.       

\subsection{MNIST : Holistic or Poetic Description as Privileged Information}
\begin{figure*}[!]
	\begin{center}
		%%%% \resizebox{<width dim>}{<height dim>}{contents to be scaled}
		\resizebox{5in}{1.5in}{\includegraphics{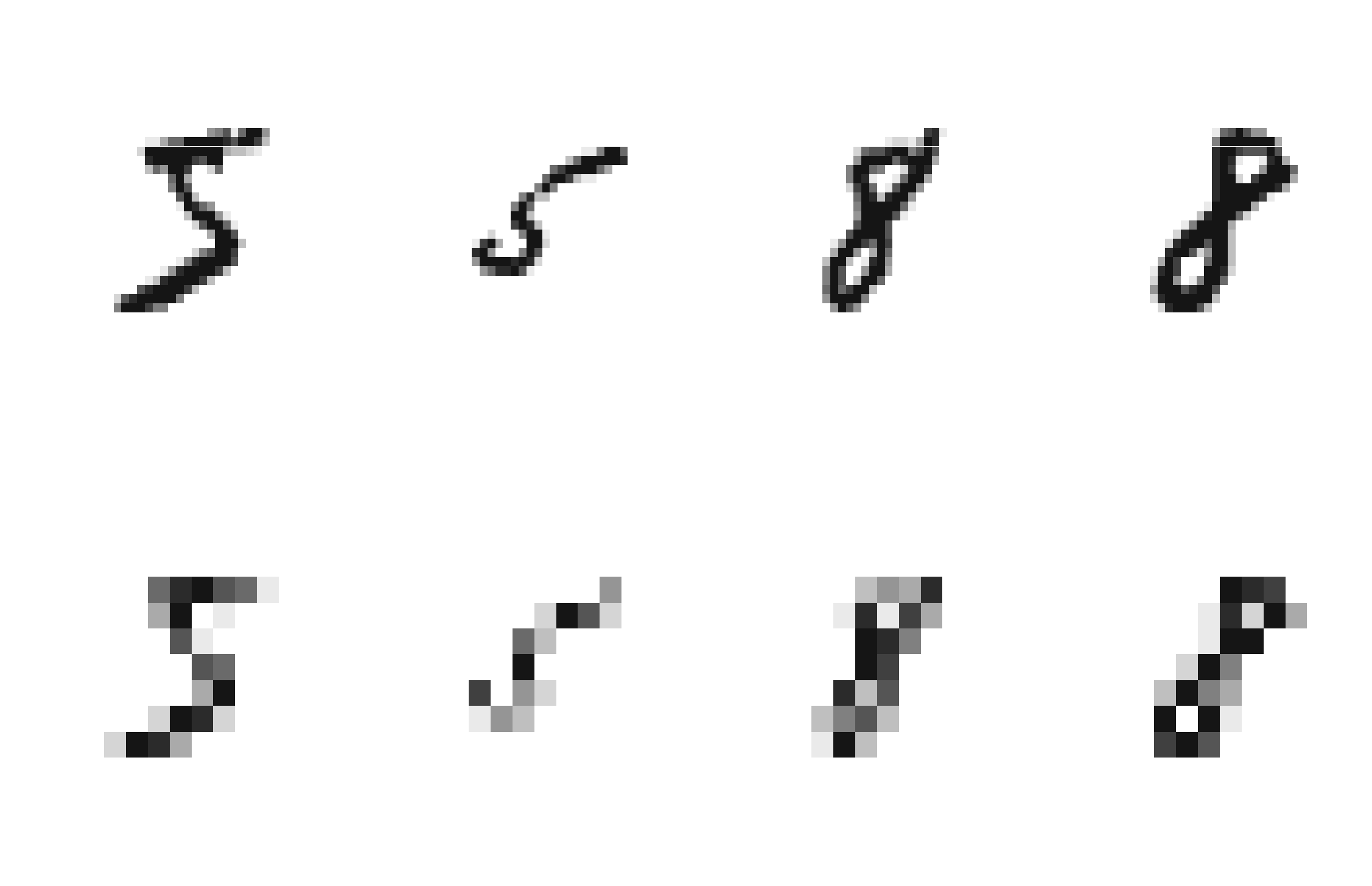}}
		\caption{Digit 5 and 8.First row shows $28\times 28$ pixel size and second row show resized image of size $10 \times 10$. }
		\label{fig:MNIST_eg}
	\end{center}
\end{figure*}

As Vapnik and Vashist \cite{vapnik2009new} select images of digit $5$ and $8$ for binary classification task, and holistic (poetic) description of these images is taken as privileged information. Similar as, we also consider the same setup of for MNIST dataset, and digit $5$ and $8$ are renamed as class $1$ and $2$, respectively. Here, images are resized from $28\times 28$ to $10 \times 10$ to make classification more difficult as mentioned in \cite{vapnik2009new}. Few examples of original and resized images are shown in Fig. \ref{fig:MNIST_eg}. Database contains 5522 and 5652 images of $5$ and $8$, respectively. As discussed above, holistic or poetic description of these training images is created \cite{vapnik2008learning} and used as privileged information. A poetic description of the first image of digit $5$ in Fig. \ref{fig:MNIST_eg} is mentioned as follows \cite{vapnik2009new}:

\begin{displayquote}
	\textit{Not absolute two-part creature. Looks more like one impulse. As for two-partness the head is a sharp tool and the bottom is round
		and flexible. As for tools it is a man with a spear ready to throw it. Or a man is shooting an arrow. He is firing the bazooka. He swung his arm, he drew back his arm and is ready to strike. He is running. He is flying. He is looking ahead. He is swift. He is throwing a spear ahead. He is dangerous. It is slanted to the right. Good snaked-ness. The snake is attacking. It is going to jump and bite. It is free and absolutely open to anything. It shows itself, no kidding. Its bottom only slightly (one point!) is on earth. He is a sportsman and in the process of training. The straight arrow and the smooth flexible body. This creature is contradictory - angular part and slightly roundish part. The lashing whip (the rope with a handle). A toe with
		a handle. It is an outside creature, not inside. Everything is finite and open. Two open pockets, two available holes, two containers. A piece of rope with a handle. Rather thick. No loops, no saltire. No hill at all. Asymmetrical. No curlings.}
\end{displayquote}

Similar as above, poetic or holistic description of first image of digit $8$ in Fig. \ref{fig:MNIST_eg} is mentioned as follows \cite{vapnik2009new}:
\begin{displayquote}
	\textit{Two-part creature. Not very perfect infinite way. It has a deadlock, a blind alley. There is a small right-hand head appendix, a small shoot. The right-hand appendix. Two parts. A bit disproportionate. Almost equal. The upper one should be a bit smaller. The starboard list is quite right. It is normal like it should be. The lower part is not very steady. This creature has a big head and too small bottom for this head. It is nice in general but not very self-assured. A rope with two loops which do not meet well. There is a small upper right-hand tail. It does not look very neat. The rope is rather good - not very old, not very thin, not very thick. It is rather like it should be. The sleeping snake which did not hide the end of its tail. The rings are not very round - oblong - rather thin oblong. It is calm. Standing. Criss-cross. The criss-cross upper angle is rather sharp. Two criss-cross angles are equal. If a tool it is a lasso. Closed absolutely. Not quite symmetrical (due to the horn).}
\end{displayquote}

% Table generated by Excel2LaTeX from sheet 'MNIST'
\begin{table}[!bp]
	\centering
	\caption{Average precision score for MNIST dataset}
	\begin{adjustbox}{width=0.7\textwidth,center}
		\begin{tabular}{|cc|ccc|c|}
			\hline
			\multicolumn{1}{|c|}{Datasets} & Group Attribute & \multicolumn{1}{c|}{OCSVM+} & \multicolumn{1}{c|}{SVDD+} & KOC   & KOC+ \bigstrut\\
			\hline
			\multicolumn{1}{|c|}{MNIST(1)} & \multirow{2}[4]{*}{Poetic Description} & \multicolumn{1}{c|}{63.6} & \multicolumn{1}{c|}{63.6} & 63.2  & \textbf{72.3} \bigstrut\\
			\cline{1-1}\cline{3-6}    \multicolumn{1}{|c|}{MNIST(2)} &       & \multicolumn{1}{c|}{79.2} & \multicolumn{1}{c|}{79.2} & 78.8  & \textbf{79.6} \bigstrut\\
			\hline
			\multicolumn{2}{|c|}{\textbf{Average =}} & 71.4  & 71.4  & 71.0  & \textbf{76.0} \bigstrut\\
			\hline
		\end{tabular}%
	\end{adjustbox}
	\label{tab:mnist_priv}%
\end{table}%

These poetic description is transformed into $21$ features like two-part-ness (0 - 5); tilting to the right (0 -
3); aggressiveness (0 - 2); stability (0 - 3); uniformity(0 - 3) etc. Here, first digit $5$ and $8$ contain features $[2, 1, 2, 0,
1]$, and $[4, 1, 1, 0, 2]$, respectively.

We have used the same training, validation and testing split\footnote{\url{https://github.com/Chandan-IITI/svmplus_matlab/tree/master/data}} as used in \cite{vapnik2009new,li2016fast}. It can be observed from Table \ref{tab:mnist_priv} that KOC+ significantly performed better for class $1$ and slightly better in case of class $2$ compared to existing methods.   

% Table generated by Excel2LaTeX from sheet 'Heart'
\begin{table}[!tp]
	\centering
	\caption{Detailed 5-Fold average precision score for Statlog heart disease dataset with different group attributes: KOC vs. KOC+}
	\begin{adjustbox}{width=0.7\textwidth,center}
    \begin{tabular}{|c|c|c|c|c|c|c|c|}
	\hline
	Datasets &       & \multicolumn{1}{l|}{Fold-1} & \multicolumn{1}{l|}{Fold-2} & \multicolumn{1}{l|}{Fold-3} & \multicolumn{1}{l|}{Fold-4} & \multicolumn{1}{l|}{Fold-5} & \multicolumn{1}{l|}{Average} \bigstrut\\
	\hline
	\multirow{9}[18]{*}{Heart(1)} &       & \multicolumn{5}{c|}{Group Attribute = Age} &  \bigstrut\\
	\cline{2-8}          & KOC   & 78.9  & 83.5  & 92.8  & 78.7  & 87.1  & 84.2 \bigstrut\\
	\cline{2-8}          & KOC+  & 90.7  & 91.1  & 90.0  & 87.9  & 90.8  & 90.1 \bigstrut\\
	\cline{2-8}          &       & \multicolumn{5}{c|}{Group Attribute = Electrocardiographic} &  \bigstrut\\
	\cline{2-8}          & KOC   & 78.0  & 78.8  & 85.8  & 76.4  & 87.5  & 81.3 \bigstrut\\
	\cline{2-8}          & KOC+  & 89.3  & 87.2  & 92.5  & 71.4  & 93.9  & 86.8 \bigstrut\\
	\cline{2-8}          &       & \multicolumn{5}{c|}{Group Attribute = Sex} &  \bigstrut\\
	\cline{2-8}          & KOC   & 78.3  & 86.5  & 92.7  & 75.8  & 86.9  & 84.0 \bigstrut\\
	\cline{2-8}          & KOC+  & 84.5  & 86.0  & 80.9  & 90.1  & 88.0  & 85.9 \bigstrut\\
	\hline
	\multirow{9}[18]{*}{Heart(2)} &       & \multicolumn{5}{c|}{Group Attribute = Age} &  \bigstrut\\
	\cline{2-8}          & KOC   & 76.5  & 84.3  & 68.1  & 69.6  & 76.6  & 75.0 \bigstrut\\
	\cline{2-8}          & KOC+  & 75.9  & 85.6  & 93.6  & 81.9  & 86.2  & 84.6 \bigstrut\\
	\cline{2-8}          &       & \multicolumn{5}{c|}{Group Attribute = Electrocardiographic} &  \bigstrut\\
	\cline{2-8}          & KOC   & 76.9  & 86.0  & 71.3  & 64.7  & 77.1  & 75.2 \bigstrut\\
	\cline{2-8}          & KOC+  & 85.0  & 77.3  & 94.4  & 84.9  & 76.3  & 83.6 \bigstrut\\
	\cline{2-8}          &       & \multicolumn{5}{c|}{Group Attribute = Sex} &  \bigstrut\\
	\cline{2-8}          & KOC   & 74.7  & 81.0  & 66.6  & 62.8  & 70.4  & 71.1 \bigstrut\\
	\cline{2-8}          & KOC+  & 74.0  & 88.3  & 93.6  & 86.3  & 78.2  & 84.1 \bigstrut\\
	\hline
\end{tabular}%
	\end{adjustbox}
	\label{tab:heart_priv_fold}%
\end{table}%

\subsection{Statlog Heart} 
We select three attributes viz., Age, Electrocardiographic, Sex, alternately from this dataset as a privileged information and transform these into group attributes. If Age$<40$ then group-1, if $40 \leq$ Age $< 50$ then group-2, and if $50 \leq$ Age $< 60$ then group-3. If Sex=$0$ then group-1, and if Sex=$1$ then group-2. If Electrocardiographic is equal to $0$ then group-1, if equal to $1$ then group-2, and if equal to $2$ then group-3. We have used 5-fold cross validation in our experiment. Results of each fold are given in detail in Table \ref{tab:heart_priv_fold}. It can be observed from Table \ref{tab:heart_priv} that KOC+ outperformed all $3$ existing one-class classifiers by significant margin.
 
% Table generated by Excel2LaTeX from sheet 'Heart'
\begin{table}[htbp]
	\centering
	\caption{Average precision score for Statlog heart disease dataset}
	\begin{adjustbox}{width=0.7\textwidth,center}
    \begin{tabular}{|cc|c|c|c|c|}
	\hline
	\multicolumn{1}{|c|}{Datasets} & \multicolumn{1}{p{9.5em}|}{Group Attribute} & OCSVM+ & SVDD+ & KOC   & \textbf{KOC+} \bigstrut\\
	\hline
	\multicolumn{1}{|c|}{\multirow{3}[6]{*}{Heart(1)}} & \multicolumn{1}{p{9.5em}|}{Age} & 84.3  & 85.4  & 84.2  & \textbf{90.1} \bigstrut\\
	\cline{2-6}    \multicolumn{1}{|c|}{} & \multicolumn{1}{p{9.5em}|}{Electrocardiographic} & 81.0  & 83.7  & 81.3  & \textbf{86.8} \bigstrut\\
	\cline{2-6}    \multicolumn{1}{|c|}{} & \multicolumn{1}{p{9.5em}|}{Sex} & 83.8  & 83.9  & 84.0  & \textbf{85.9} \bigstrut\\
	\hline
	\multicolumn{1}{|c|}{\multirow{3}[6]{*}{Heart(2)}} & \multicolumn{1}{p{9.5em}|}{Age} & 81.2  & 81.6  & 75.0  & \textbf{84.6} \bigstrut\\
	\cline{2-6}    \multicolumn{1}{|c|}{} & \multicolumn{1}{p{9.5em}|}{Electrocardiographic} & 80.0  & 80.0  & 75.2  & \textbf{83.6} \bigstrut\\
	\cline{2-6}    \multicolumn{1}{|c|}{} & \multicolumn{1}{p{9.5em}|}{Sex} & 77.4  & 78.4  & 71.1  & \textbf{84.1} \bigstrut\\
	\hline
	\multicolumn{2}{|c|}{\textbf{Average =}} & 81.3  & 82.2  & 78.5  & \textbf{85.9} \bigstrut\\
	\hline
\end{tabular}%
	\end{adjustbox}
	\label{tab:heart_priv}%
\end{table}%

% Table generated by Excel2LaTeX from sheet 'Abalone'
\begin{table}[htbp]
	\centering
	\caption{Detailed 5-Fold  average precision score for Abalone dataset with different group attributes: KOC vs. KOC+}
	\begin{adjustbox}{width=0.7\textwidth,center}
    \begin{tabular}{|c|c|c|c|c|c|c|c|}
	\hline
	Datasets &       & \multicolumn{1}{l|}{Fold-1} & \multicolumn{1}{l|}{Fold-2} & \multicolumn{1}{l|}{Fold-3} & \multicolumn{1}{l|}{Fold-4} & \multicolumn{1}{l|}{Fold-5} & \multicolumn{1}{l|}{Average} \bigstrut\\
	\hline
	\multirow{9}[18]{*}{Abalone(1)} &       & \multicolumn{5}{c|}{Group Attribute = Height} &  \bigstrut\\
	\cline{2-8}          & KOC   & 77.4  & 80.4  & 61.6  & 88.6  & 88.1  & 79.2 \bigstrut\\
	\cline{2-8}          & KOC+  & 79.0  & 86.7  & 76.3  & 92.9  & 90.9  & 85.2 \bigstrut\\
	\cline{2-8}          &       & \multicolumn{5}{c|}{Group Attribute = Length} &  \bigstrut\\
	\cline{2-8}          & KOC   & 80.2  & 78.4  & 74.1  & 86.6  & 91.3  & 82.1 \bigstrut\\
	\cline{2-8}          & KOC+  & 79.6  & 86.9  & 77.6  & 92.3  & 91.8  & 85.7 \bigstrut\\
	\cline{2-8}          &       & \multicolumn{5}{c|}{Group Attribute = Whole weight} &  \bigstrut\\
	\cline{2-8}          & KOC   & 76.9  & 79.1  & 73.4  & 78.6  & 90.6  & 79.7 \bigstrut\\
	\cline{2-8}          & KOC+  & 77.7  & 86.8  & 77.6  & 92.7  & 92.6  & 85.5 \bigstrut\\
	\hline
	\multirow{9}[18]{*}{Abalone(2)} &       & \multicolumn{5}{c|}{Group Attribute = Height} &  \bigstrut\\
	\cline{2-8}          & KOC   & 34.1  & 47.6  & 33.5  & 38.5  & 38.4  & 38.4 \bigstrut\\
	\cline{2-8}          & KOC+  & 50.1  & 42.4  & 36.2  & 49.4  & 35.4  & 42.7 \bigstrut\\
	\cline{2-8}          &       & \multicolumn{5}{c|}{Group Attribute = Length} &  \bigstrut\\
	\cline{2-8}          & KOC   & 35.9  & 47.4  & 45.4  & 37.0  & 43.9  & 41.9 \bigstrut\\
	\cline{2-8}          & KOC+  & 44.5  & 37.2  & 33.8  & 33.7  & 39.2  & 37.7 \bigstrut\\
	\cline{2-8}          &       & \multicolumn{5}{c|}{Group Attribute = Whole weight} &  \bigstrut\\
	\cline{2-8}          & KOC   & 32.3  & 49.1  & 46.1  & 37.0  & 53.1  & 43.5 \bigstrut\\
	\cline{2-8}          & KOC+  & 47.3  & 34.7  & 34.5  & 39.7  & 42.5  & 39.7 \bigstrut\\
	\hline
	\multirow{9}[18]{*}{Abalone(3)} &       & \multicolumn{5}{c|}{Group Attribute = Height} &  \bigstrut\\
	\cline{2-8}          & KOC   & 53.9  & 52.4  & 51.3  & 47.6  & 58.2  & 52.7 \bigstrut\\
	\cline{2-8}          & KOC+  & \multicolumn{1}{r|}{72.4} & \multicolumn{1}{r|}{65.9} & \multicolumn{1}{r|}{71.2} & \multicolumn{1}{r|}{64.1} & \multicolumn{1}{r|}{43.8} & 63.5 \bigstrut\\
	\cline{2-8}          &       & \multicolumn{5}{c|}{Group Attribute = Length} &  \bigstrut\\
	\cline{2-8}          & KOC   & 57.1  & 49.6  & 44.3  & 48.7  & 60.0  & 51.9 \bigstrut\\
	\cline{2-8}          & KOC+  & 74.0  & 64.6  & 63.7  & 69.6  & 67.8  & 67.9 \bigstrut\\
	\cline{2-8}          &       & \multicolumn{5}{c|}{Group Attribute = Whole weight} &  \bigstrut\\
	\cline{2-8}          & KOC   & 38.3  & 47.9  & 52.0  & 49.8  & 58.8  & 49.4 \bigstrut\\
	\cline{2-8}          & KOC+  & 71.6  & 72.8  & 65.8  & 47.2  & 63.5  & 64.2 \bigstrut\\
	\hline
\end{tabular}%
	\end{adjustbox}
	\label{tab:abalone_priv_fold}%
\end{table}%

\subsection{Abalone}
In this dataset, $3$ attributes viz., Height, Length and Whole weight, are used as the privileged information one by one. First, we divide data into groups based on the values of attributes i.e. if Height$<0.15$ then group-1 otherwise group-2, if Length<$0.5$ then group-1 otherwise group-2, if Whole weight$<0.8$ then group-1 otherwise group-2. We have used 5-fold cross validation in our experiment. Results of each fold are given in detail in Table \ref{tab:abalone_priv_fold}. It can be observed from Table \ref{tab:abalone_priv} that KOC+ has exhbited significant performance improvement over existing ones in most of the cases.   

% Table generated by Excel2LaTeX from sheet 'Abalone'
\begin{table}[!tp]
	\centering
	\caption{Average precision score for Abalone dataset}
	\begin{adjustbox}{width=0.7\textwidth,center}
    \begin{tabular}{|cc|c|c|c|c|}
	\hline
	\multicolumn{1}{|c|}{Datasets} & Group Attribute & OCSVM+ & SVDD+ & KOC   & KOC+ \bigstrut\\
	\hline
	\multicolumn{1}{|c|}{\multirow{3}[6]{*}{Abalone(1)}} & Height & 73.9  & 74.4  & 79.2  & \textbf{85.2} \bigstrut\\
	\cline{2-6}    \multicolumn{1}{|c|}{} & Length & 80.9  & 80.6  & 82.1  & \textbf{85.7} \bigstrut\\
	\cline{2-6}    \multicolumn{1}{|c|}{} & Whole weight & 78.2  & 77.9  & 79.7  & \textbf{85.5} \bigstrut\\
	\hline
	\multicolumn{1}{|c|}{\multirow{3}[6]{*}{Abalone(2)}} & Height & 52.1  & 52.3  & 38.4  & 42.7 \bigstrut\\
	\cline{2-6}    \multicolumn{1}{|c|}{} & Length & \textbf{54.5} & 52.6  & 41.9  & 37.7 \bigstrut\\
	\cline{2-6}    \multicolumn{1}{|c|}{} & Whole weight & 56.3  & \textbf{56.7} & 43.5  & 39.7 \bigstrut\\
	\hline
	\multicolumn{1}{|c|}{\multirow{3}[6]{*}{Abalone(3)}} & Height & 50.8  & 51.3  & 52.7  & \textbf{63.5} \bigstrut\\
	\cline{2-6}    \multicolumn{1}{|c|}{} & Length & 46.1  & 46.1  & 51.9  & \textbf{67.9} \bigstrut\\
	\cline{2-6}    \multicolumn{1}{|c|}{} & Whole weight & 48.5  & 47.3  & 49.4  & \textbf{64.2} \bigstrut\\
	\hline
	\multicolumn{2}{|c|}{\textbf{Average =}} & 60.1  & 59.9  & 57.7  & 63.6 \bigstrut\\
	\hline
\end{tabular}%
	\end{adjustbox}
	\label{tab:abalone_priv}%
\end{table}%

\section{Conclusion}\label{sec:concl}
We have modified the formulation of KOC and enabled it to learn from the privileged information. Experimental results have exhibited that how privileged information helps in the performance improvement over the tradition KRR-based one-class classifier (KOC). On the one hand, privileged information is useless for the minimization problem of KOC, however, on the other hand, structure of minimization problem of KOC+ can handle it effectively by introducing a correction function. Moreover, results are better than existing privileged information-based methods OCSVM+ and SVDD+, as well as, KOC+ obtains solution by a non-iterative approach, therefore, it will consume less training time compared to iterative approach based methods viz., OCSVM+ and SVDD+. There are various possible real world applications of this approach, especially in the field of intrusion detection, medical diagnosis etc., which needs to explore further.    

\section*{Acknowledgment}
This research was supported by Department of Electronics and Information Technology (DeITY, Govt. of India) under Visvesvaraya PhD scheme for electronics \& IT.

\vskip5pt

\bibliographystyle{unsrt}

\begin{thebibliography}{10}

\bibitem{wang2018learning}
S.~Wang, S.~Chen, T.~Chen, and X.~Shi.
\newblock Learning with privileged information for multi-label classification.
\newblock {\em Pattern Recognition}, 81:60--70, 2018.

\bibitem{chevalier2018classifying}
M.~Chevalier, N.~Thome, G.~H{\'e}naff, and M.~Cord.
\newblock Classifying low-resolution images by integrating privileged
  information in deep cnns.
\newblock {\em Pattern Recognition Letters}, 116:29--35, 2018.

\bibitem{lambert2018deep}
J.~Lambert, O.~Sener, and S.~Savarese.
\newblock Deep learning under privileged information using heteroscedastic
  dropout.
\newblock In {\em Proceedings of the IEEE Conference on Computer Vision and
  Pattern Recognition}, pages 8886--8895, 2018.

\bibitem{smolyakov2016one}
E.~Burnaev and D.~Smolyakov.
\newblock One-class svm with privileged information and its application to
  malware detection.
\newblock In {\em IEEE International Conference on Data Mining Workshops, ICDM
  Workshops}, pages 12--15, 2016.

\bibitem{motiian2016information}
S.~Motiian, M.~Piccirilli, Donald~A. Adjeroh, and G.~Doretto.
\newblock Information bottleneck learning using privileged information for
  visual recognition.
\newblock In {\em Proceedings of the IEEE Conference on Computer Vision and
  Pattern Recognition}, pages 1496--1505, 2016.

\bibitem{zhang2015support}
W.~Zhang.
\newblock Support vector data description using privileged information.
\newblock {\em Electronics Letters}, 51(14):1075--1076, 2015.

\bibitem{vapnik2015learning}
V.~Vapnik and R.~Izmailov.
\newblock Learning using privileged information: similarity control and
  knowledge transfer.
\newblock {\em Journal of machine learning research}, 16(2023-2049):2, 2015.

\bibitem{lapin2014learning}
M.~Lapin, M.~Hein, and B.~Schiele.
\newblock Learning using privileged information: Svm+ and weighted svm.
\newblock {\em Neural Networks}, 53:95--108, 2014.

\bibitem{zhu2014new}
W.~Zhu and P.~Zhong.
\newblock A new one-class svm based on hidden information.
\newblock {\em Knowledge-Based Systems}, 60:35--43, 2014.

\bibitem{feyereisl2012privileged}
Jan Feyereisl and Uwe Aickelin.
\newblock Privileged information for data clustering.
\newblock {\em Information Sciences}, 194:4--23, 2012.

\bibitem{vapnik2009new}
V.~Vapnik and A.~Vashist.
\newblock A new learning paradigm: Learning using privileged information.
\newblock {\em Neural networks}, 22(5-6):544--557, 2009.

\bibitem{xu2015distance}
X.~Xu, W.~Li, and D.~Xu.
\newblock Distance metric learning using privileged information for face
  verification and person re-identification.
\newblock {\em IEEE transactions on neural networks and learning systems},
  26(12):3150--3162, 2015.

\bibitem{scholkopf1999support}
B.~Sch{\"o}lkopf, R.~C. Williamson, A.~J. Smola, J.~Shawe-Taylor, and J.~C.
  Platt.
\newblock Support vector method for novelty detection.
\newblock In {\em NIPS}, volume~12, pages 582--588, 1999.

\bibitem{tax1999support}
D.~M.~J. Tax and R.~P.~W. Duin.
\newblock Support vector domain description.
\newblock {\em Pattern recognition letters}, 20(11):1191--1199, 1999.

\bibitem{leng2014one}
Q.~Leng, H.~Qi, J.~Miao, W.~Zhu, and G.~Su.
\newblock One-class classification with extreme learning machine.
\newblock {\em Mathematical Problems in Engineering}, pages 1--11, 2014.

\bibitem{Dua:2017}
D.~Dheeru and E.~Karra~Taniskidou.
\newblock {UCI} machine learning repository, 2017.

\bibitem{sharmanska2013learning}
V.~Sharmanska, N.~Quadrianto, and C.~H. Lampert.
\newblock Learning to rank using privileged information.
\newblock In {\em Proceedings of the IEEE International Conference on Computer
  Vision}, pages 825--832, 2013.

\bibitem{argyriou2009there}
A.~Argyriou, C.~A. Micchelli, and M.~Pontil.
\newblock When is there a representer theorem? vector versus matrix
  regularizers.
\newblock {\em Journal of Machine Learning Research}, 10(Nov):2507--2529, 2009.

\bibitem{vapnik2008learning}
V.~Vapnik, A.~Vashist, and N.~Pavlovitch.
\newblock Learning using hidden information: Master class learning.
\newblock {\em NATO Science for Peace and Security Series, D: Information and
  Communication Security}, 19:3--14, 2008.

\bibitem{li2016fast}
W.~Li, D.~Dai, M.~Tan, D.~Xu, and L.~V.~Gool.
\newblock Fast algorithms for linear and kernel svm+.
\newblock In {\em Proceedings of the IEEE Conference on Computer Vision and
  Pattern Recognition}, pages 2258--2266, 2016.

\end{thebibliography}

\end{document}